\documentclass[letterpaper, 10 pt, conference]{ieeeconf}  

\IEEEoverridecommandlockouts                              

\usepackage[linesnumbered,ruled,vlined]{algorithm2e}
\usepackage{algorithmic}
\usepackage{amsmath,amsfonts,amssymb}
\usepackage{graphicx}
\graphicspath{{figures/}}
\usepackage{epsfig} 
\usepackage{times}
\usepackage{siunitx}
\usepackage{hyperref}
\usepackage{booktabs}
\usepackage[capitalise]{cleveref}
\let\labelindent\relax
\usepackage[inline]{enumitem}
\usepackage{multirow}
\usepackage[normalem]{ulem}
\Crefname{algorithm}{Alg.}{Algs.}
\Crefname{section}{Sec.}{Secs.}
\Crefname{equation}{Eq.}{Eqs.}
\usepackage{caption}
\usepackage{authblk}
\usepackage{booktabs}
\usepackage{comment}

\usepackage[
    sorting=none,
    giveninits=true,
    maxbibnames=9,
    maxcitenames=2,backend=biber
    ]{biblatex}
\renewcommand{\baselinestretch}{0.999}

\newcommand{\kstef}{k_\mathrm{stef}}

\title{\LARGE \bf
Fast Online Learning of CLiFF-maps in Changing Environments}

\author{
Yufei Zhu$^{1,2}$, Andrey Rudenko$^{2}$, Luigi Palmieri$^{2}$, Lukas Heuer$^{1,2}$, Achim J. Lilienthal$^{1,3}$, Martin Magnusson$^{1}$%
\thanks{$^{1}$Robot Navigation and Perception Lab, AASS Research Center, 
{\"O}rebro University, Sweden {\tt\small yufei.zhu@oru.se}}%
\thanks{$^{2}$Robert Bosch GmbH, Corporate Research, Stuttgart, Germany}%
\thanks{$^{3}$Chair: Perception for Intelligent Systems, Technical University of Munich, Germany}%
\thanks{This work has received funding from the European Union’s Horizon 2020 research and innovation programme under grant agreement No 101017274 (DARKO).}
}

\begin{document}

\maketitle
\thispagestyle{empty}
\pagestyle{empty}

\begin{abstract}
Maps of dynamics are effective representations of motion patterns learned from prior observations, with recent research demonstrating their ability to enhance various downstream tasks such as human-aware robot navigation, long-term human motion prediction, and robot localization. Current advancements have primarily concentrated on methods for learning maps of human flow in environments where the flow is static, i.e., not assumed to change over time. 
In this paper we propose an online update method of the CLiFF-map (an advanced map of dynamics type that models motion patterns as velocity and orientation mixtures) to actively detect and adapt to human flow changes. As new observations are collected, our goal is to update a CLiFF-map to effectively and accurately integrate them, while retaining relevant historic motion patterns. The proposed online update method maintains a probabilistic representation in each observed location, updating parameters by continuously tracking sufficient statistics. In experiments using both synthetic and real-world datasets, we show that our method is able to maintain accurate representations of human motion dynamics, contributing to high performance flow-compliant planning downstream tasks, while being orders of magnitude faster than the comparable baselines.
\end{abstract}



\section{Introduction} \label{section-introduction}

Safe and efficient operation in complex, dynamic and densely crowded human environments is a critical prerequisite for deploying robots in various tasks to support people in their daily activities \cite{kruse2013human,triebel2016spencer,mavrogiannis2023core}. Considerable efforts are dedicated to support the deployment of mobile robots by enabling them to follow social norms and achieve legible and socially compliant navigation \cite{lasota2017survey}.

Extending the environment model with human motion patterns using a \emph{map of dynamics} (MoD) is one way to achieve unobtrusive navigation compliant with existing motion flows in the environment \cite{palmieri2017kinodynamic,swaminathan-2022-benchmarking}, or avoid crowded areas \cite{doellinger2018predicting}.
MoDs are an increasingly popular tool to store spatial-temporal information about patterns of motion in an environment, such as the motions of humans \cite{tomasz_survey23}. Maps of dynamics extend the geometric world model, improving the temporal and contextual awareness of the robot's own configuration space. These maps are useful for several downstream tasks, e.g. to predict long-term motion trajectories \cite{zhu2023clifflhmp,almeida23}, improve localization in dynamic environments \cite{krajnik2017fremen} and robot navigation in crowded spaces \cite{molina2023iliad}, particularly planning safe and unobtrusive trajectories~\cite{palmieri2017kinodynamic,swaminathan2018down} and efficiently allocating complex tasks \cite{liu2023human}. In particular, the CLiFF-map (circular-linear flow field)~\cite{kucner2017enabling} is a powerful and flexible representation to store 
velocity, orientation and turbulence of the multi-modal human motion flows in the environment, implicitly incorporating such features as common goal locations, restricted areas and other semantic attributes \cite{schreiter2024thormagni}. Practically implemented on a robot, CLiFF-map generalizes prior observations in a compact spatial representation that can be efficiently queried by downstream components.

However, prior art typically assumes that once learned, the map of dynamics remains constant~\cite{kucner2020probabilistic}, or follows fixed periodic patterns every day~\cite{molina2018modelling}. In reality, as the robot operates continuously, the motion patterns may change due to various events or alterations in the environment topology and semantics \cite{liu2023human}. The map of dynamics should actively detect and accommodate these changes. This motivates the need for an online update method, similar to the one developed for iterative improvement of periodic maps of dynamics \cite{molina22exploration}. Furthermore, as the robot operates over extended periods, the accumulation of human observation data makes rebuilding the map from scratch increasingly costly. An online learning model becomes practical and necessary in scenarios where handling large datasets is infeasible due to storage limitations or privacy concerns. This situation motivates the need for incremental and efficient map updates, which augment previously accumulated knowledge with recent observations without retaining all past dynamics samples.

\begin{figure}
\centering
\includegraphics[width=1.\linewidth]{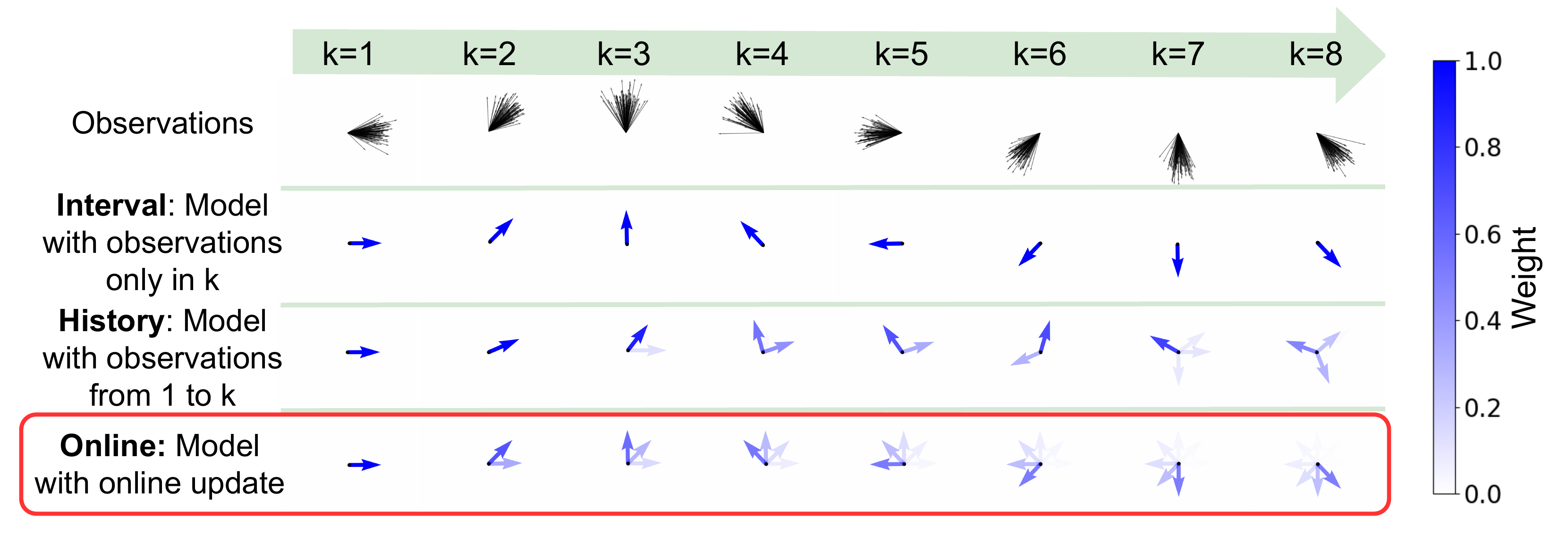}
\caption{Online update results of toy example data compared with using only new observations and using all observations to build the model. The \textbf{top row} shows raw observations for each of the eight directions $(0^\circ, 45^\circ, 90^\circ, 135^\circ, 180^\circ, 225^\circ, 270^\circ, 315^\circ)$, provided in each iteration $k$. Blue arrows depict the mean vectors of a CLiFF Gaussian mixture model, with transparency indicating the component weights. Three modeling approaches are compared: the \textbf{second row} shows models built using only observations from the current iteration $k$; the \textbf{third row} shows models built with cumulative observations from iteration 1 to $k$, which are overgeneralized and fail to prioritize recent observations; the \textbf{fourth row} shows the proposed online update method, which incorporate new data while retaining relevant historical patterns, offering a dynamic representation of the motion pattern over time.}
\label{fig:art_demo}
\vspace*{-2mm}
\end{figure}

In this paper we propose a method for online updating of the CLiFF-map of dynamics for mobile robots using a variation of the stochastic expectation maximization algorithm. As new observations are collected, our goal is to update the existing representation to effectively and accurately integrate the new information. At the same time, the robot should not immediately dismiss previously learned patterns while avoiding the need to store the entire historical dataset. The proposed online update method maintains the probabilistic representation in each observed location, updating parameters by continuously tracking sufficient statistics. 
As shown in \cref{fig:art_demo}, our method not only ensures that the model remains adaptively accurate in reflecting the most recent human motion but also maintains consistency with historical data, thereby preserving a comprehensive understanding of the environment over time. In experiments on both a synthetic dataset and the real-world ATC \cite{brscic2013person} dataset, we show that our method quickly recognizes changes in environments with sparse and dense motion flows, while being significantly faster than baseline methods. Qualitative results further show that the maps learned with our online update method represent recent trajectories more accurately, and are therefore better suited for informing planning algorithms.

\section{Related Work} \label{section-relatedwork}
There are several types of Maps of Dynamics (MoDs) described in the literature, generally striving to provide an efficient tool for storing and querying information about historical or expected changes in states within the environment. Some MoDs model the expected state of occupancy or other binary signals~\cite{krajnik2017fremen, saarinen-2012-imac}, or recurring discrete states of objects~\cite{burgard2007mobile}. MoDs can be built from various sources of input, such as trajectories~\cite{bennewitz2005learning}, dynamics samples, or information about the flow of continuous media (e.g., air or water)~\cite{bennetts-2017-probabilistic}. Furthermore, these models can feature diverse underlying representations, including evidence grids, histograms~\cite{zhu2024lace}, graphs, or Gaussian mixtures. We refer the reader to \cite{tomasz_survey23} for a comprehensive survey. In this section, we will focus on MoDs that are designed for, or have been applied specifically to, the motion of people (trajectories, directions, velocities).


Bennewitz et al.~\cite{bennewitz2005learning} learn a collection of human motion patterns from clustered trajectory data (with $k$-means clustering), using Gaussian mixtures and expectation maximization (EM).
A drawback of this method is that such clustering does not lend itself well to online updates with new data and relies on having complete observations of full trajectories in order to cluster the motion patterns that make up the map. Doellinger et al.~\cite{doellinger2019environment} employ a convolutional neural network trained on simulated trajectories to predict where people are likely to walk, based on an input occupancy grid map. While it does not require full trajectories for training, it cannot easily integrate new observations to update the model. Similarly, \cite{verdoja2022generating} predicts a map of dynamics given a static occupancy map. However, the output map in this case is a Bayesian floor field, where each point in the map stores the probabilities of observing motion along a discrete set of directions, rather than only the probability of occupancy.

The CLiFF-map representation~\cite{kucner2017enabling} has been used recently to facilitate efficient human-aware motion planning~\cite{palmieri2017kinodynamic,swaminathan2018down,swaminathan-2022-benchmarking} and long-term human motion prediction~\cite{zhu2023data,zhu2023clifflhmp,almeida2024performance}. CLiFF-map represents local flow patterns as a multi-modal, continuous joint distribution of speed and orientation, as further described in Section~\ref{section-method}. A benefit of CLiFF-map is that it can be built from incomplete or spatially sparse velocity observations \cite{almeida2024performance}, without the need to store a long history of data or deploy advanced tracking algorithms. However, like most other types of existing MoDs, including those mentioned above, CLiFF-maps are typically built offline. A key reason for this is the high computational costs associated with the building process. This constraint limits their applicability in real environments, where patterns can change over time, e.g. due to changes in environment topology and semantics.

STeF-map~\cite{molina2018modelling} is another representation that builds a local motion model per cell of a grid map. STeF-map uses a discrete set of eight directions, maintaining a model of when motion can be observed at that point in that direction to learn periodic flow patterns. As a periodic spatiotemporal map of dynamics, it can be used to predict activities at specific times of the day, under the assumption that the periodic patterns exist in this environment. Most closely related to the present work is \cite{molina22exploration}, which uses incremental online updates to iteratively improve the periodic patterns in the STeF-map. In contrast, we assume that the map may be non-stationary (e.g., due to changing obstacle configurations or temporary external conditions \cite{schreiter2024thormagni}), and our online update is designed to adapt to such changes quickly, while not forgetting past flow patterns.

To enable incremental online updates of the CLiFF-map,  we use a stochastic expectation maximization algorithm~\cite{Capp09onlineEM} to process streaming data, as further described in Section~\ref{section-method}, and to achieve an accurate representation of dynamics, gradually balancing historical observations with more recent samples, as demonstrated in Section~\ref{section-results}. 






\section{Online CLiFF-map Update Method} \label{section-method}

\subsection{Problem statement}
A given geometric environment is denoted as \( M \subseteq \mathbb{R}^2 \), which is discretized to a set of locations $L=\{\mathrm{loc}_1, \mathrm{loc}_2, ..., \mathrm{loc}_{|L|}\}$. Each location $\mathrm{loc}_l \in L$ aggregates observations collected within a radius $r$ of $\mathrm{loc}_l$. Our aim is to learn a dynamics model at each location that effectively and accurately represents non-stationary local motion patterns. As new observations become available, the model will be dynamically updated to integrate the information from them.

\newcommand{\vect}[1]{\mathbf{#1}} 
\newcommand{\observation}{\vect{y}} 
\newcommand{\direction}{\theta}
\newcommand{\speed}{\rho}
\newcommand{\mean}{\boldsymbol{\mu}}
\newcommand{\cov}{\boldsymbol{\Sigma}}
\newcommand{\SWND}{\mathcal{N}^\mathrm{SW}}
\newcommand{\mix}{m} 

\subsection{Definitions}

CLiFF-map represents motion patterns using multimodal statistics to represent speed and orientation jointly~\cite{kucner2017enabling}. 
In the CLiFF-map, each location is associated with a Semi-Wrapped Gaussian Mixture Model (SWGMM)~\cite{Roy16SWGMM} to capture the dependency between the speed and orientation, representing motion patterns based on local observations.

The SWGMM represents speed and direction jointly as velocity $\mathbf{V} = [\theta, \rho]^\top$ using direction $\theta$ and speed $\rho$, where $\rho \in \mathbb{R}^+$, $\theta \in [0,2\pi)$. This semi-wrapped probability density function (PDF) over velocities can be visualized as a function on a cylinder. Direction values $\theta$ are wrapped on the unit circle and the speed $\rho$ runs along the length of the cylinder.

Since the direction $\theta$ is a circular variable and the speed is linear, a mixture of \emph{semi-wrapped} normal distributions (SWNDs) is used. Encoding the multimodal characteristic of human motion flow, an SWGMM is a PDF represented as a weighted sum of $J$ SWNDs: 

\begin{equation}
p(\mathbf{V} | \mathbf{\xi}) = \sum_{j=1}^{J}\mix_{j}\SWND_{\cov_j,\mean_j}(\mathbf{V}), 
\end{equation}

\noindent where $\boldsymbol{\xi} = \{ \xi_j = (\mix_j, \mean_j, \cov_j) | j \in \mathbb{Z}^+ \}$ denotes a finite set of components of the SWGMM, $\mix_j$ denotes the mixing factor and satisfies $0 \leq \mix_j \leq 1$, and an SWND $\SWND_{\cov, \mean}$ is formally defined as 
\begin{equation}
\SWND_{\cov, \mean}(\mathbf{V}) = \sum_{w \in \mathbb{Z}} \mathcal{N}_{\cov, \mean}( [\theta,  \rho]^\top + 2\pi [ w,  0 ]^\top ),
\end{equation}

\noindent where $\cov, \mean$ denote the covariance matrix and mean value of the directional velocity $(\theta, \rho)^\top$, and $w$ is a winding number. Although $w \in \mathbb{Z}$, the PDF can be approximated adequately by taking $w \in \{-1, 0, 1\}$ for practical purposes \cite{jupp2018circlulalrbook}.

When the first batch of observations for location $\mathrm{loc}_l$ is available, SWGMM parameters are estimated using expectation maximization (EM)~\cite{Dempster77iEM} along with mean shift~\cite{cheng95meanshift} to obtain the number and initial positions of dynamics modes. When collecting new observations at the same location $\mathrm{loc}_l$, SWGMM parameters are updated using the stochastic EM method (sEM~\cite{Capp09onlineEM}), which is a fast online variant of the EM algorithm. In sEM, the expectation step of the original EM algorithm is replaced by a stochastic approximation step, while the maximization step remains unchanged.

\begin{algorithm}[t]
\small
    \KwIn{Number of observing iterations $K$}
    \KwOut{online-CLiFF-map $\Xi_{\mathrm{loc}_1,..,\mathrm{loc}_{|L|}}$}
        \For { $k = 1, ..., K$} {
            \For {$\mathrm{loc}_l = \mathrm{loc}_1, ..., \mathrm{loc}_{|L|}$} {
                $\observation_k \leftarrow $ getObservation($\mathrm{loc}_l$) \\
                \eIf { $\Xi_{\mathrm{loc}_l} = \varnothing$} {
                    $\Xi_{\mathrm{loc}_l} \leftarrow $ buildCLiFFmap($\observation_k$) \\
                }{
                    $\hat{s}_k \leftarrow $ sE--step \\ 
                    $\mix_{1,..,J}$, $\mean_{1,..,J}$, $\cov_{1,..,J} \leftarrow $ M--step \\ 
                }
            }
        }
    \Return $\Xi_{\mathrm{loc}_1,..,\mathrm{loc}_{|L|}}$
\caption{Online MoD update}
\label{alg:onlineMoDalgo}
\end{algorithm}

\subsection{Online map update algorithm}
The online update method is summarized in \cref{alg:onlineMoDalgo}. 
During each observation iteration, the online CLiFF-map is built and updated with new observations. Let $k\geq0$ be the iteration number for collecting observations. During iteration $k$, at location $\mathrm{loc}_l$, new observations $\observation_k$ are collected over the time interval $[t, t + \Delta t]$ (line 3 of \cref{alg:onlineMoDalgo}). If no model has previously been built at $\mathrm{loc}_l$, i.e., no motion has been observed at $\mathrm{loc}_l$ yet, an SWGMM will be initiated with $\observation_k$ (line 5 of \cref{alg:onlineMoDalgo}), where $\observation_k$ denotes the set of velocity vectors observed in iteration $k$ of human motion. Otherwise, $\observation_k$ will be used to update the existing model for $\mathrm{loc}_l$. The variable $N_k$ represents the number of new observations collected at $\mathrm{loc}_l$ during the interval. Let $\hat{s}_{k}$ be the estimated sufficient statistics at iteration $k$, for each component $j$, where $\hat{s}_{j}$ is composed of $(s^{(1)}_{j}, s^{(2)}_{j}, s^{(3)}_{j})$ and the initial value $\hat{s}_{0}$ computed using \cref{eq5} and initial observations. Sufficient statistics contain all the information needed to estimate model parameters. The sEM method tracks the sufficient statistics using a stochastic approximation procedure. The SWGMM parameters, $\mix$, $\mean$ and $\cov$ are then updated using sEM, which consists of two sub-steps:
\begin{equation}
\textbf{sE-step: } \hat{s}_{k} =  \hat{s}_{k-1} + \gamma_{k}(S_{k} - \hat{s}_{k-1}),
\label{eq3}
\end{equation}
\begin{align}
\textbf{M-step: } \quad
    & \mix_{j} = \hat{s}^{(1)}_{j}, \\
    & \mean_j = (\hat{s}^{(1)}_{j})^{-1}\hat{s}^{(2)}_{j}, \\
    & \cov_j = (\hat{s}^{(1)}_{j})^{-1}\hat{s}^{(3)}_{j} - \mean_j\mean_j^\top.
\label{eq4}
\end{align}

In the sE--step, the sufficient statistics of SWGMM are computed and updated with a sequence of stepsizes $(\gamma_{k})_{k\geq1}$. For iteration $k$ and SWGMM components $j\in[1,J]$, the sufficient statistics $S_{k}$ is composed of $(S^{(1)}_{j}, S^{(2)}_{j}, S^{(3)}_{j})$. For partitions of the sufficient statistics $p\in[1,3]$, we have $S_j^{(p)}=N_k^{-1}\Sigma_{i=1}^{N_k}S_{i,j}^{(p)}$. 
For each observation $i$, the observed data is $\observation_i$, $S_{i,j}$ is
\begin{equation}
    S^{(1)}_{i, j} = \eta_{i,j}, 
    \quad S^{(2)}_{i, j} = \eta_{i,j}\observation_i, 
    \quad S^{(3)}_{i, j} = \eta_{i,j}\observation_{i} \observation_{i}^\top,
\label{eq5}
\end{equation}
where the responsibility $\eta$ is
\begin{equation}
    \eta_{i,j} = \frac{\mix_{j}\SWND_{\cov_j,\mean_j}(\observation_{i})}{\Sigma_{j=1}^{J}\mix_{j}\SWND_{\cov_j,\mean_j}(\observation_{i})}
    .
\end{equation}

In each iteration $k$, the step size $\gamma_k$ determines the impact of new observations on the model. To effectively update the model, we consider two factors: the motion intensity and the time elapsed since observation. Observations that occurred a long time ago are considered less important than more recent ones. To incorporate them, we use a dynamically decreasing total number of observations as an indicator, $N_{\mathrm{ind}}^k = \lambda N_{\mathrm{ind}}^{k-1}+N_k$, where $0<\lambda<1$ is the \emph{decay rate}. The step size $\gamma_k$ is defined as the ratio of the size of the new observations to $N_{\mathrm{ind}}^k$, $\gamma_{k}=N_{k}/N_{\mathrm{ind}}^k$. This ensures the influence of each new observation diminishes over time as more observations are accumulated, and more recent data have a proportionally higher influence than older ones. This dynamic adjustment helps the model gradually incorporate new information while reducing the weight of the older data points.
During the online update, if the new observations do not fit the current model, the number of SWGMM components is increased based on the condition $\sum_{j=1}^{J}\sum_{i=1}^{N_k}\eta_{i,j} < \eta_{\mathrm{thres}}$, where $\eta_{\mathrm{thres}}$ is a predefined threshold. A smaller $\eta_{\mathrm{thres}}$ makes it easier to classify new observations as not fitting well with the current model, thereby triggering an adjustment in the number of SWGMM components. The updated SWGMM then consists of both previous and new SWNDs, built using recent observations. Their weights are scaled by $\gamma_k$ for new SWNDs and $1-\gamma_k$ for previous ones. All weights are then normalized to integrate the historical and new components within the model.

\section{Experiments} \label{section-experiments}

\subsection{Datasets}
We evaluate the online CLiFF method and a set of baselines on both real-world and synthetic datasets. To reflect the iterative nature of building the maps of dynamics from older and more recent observations, we divide each dataset into batches of trajectory data, corresponding to the input iterations as observed by the robot. The aim is to show how well the most recent map of dynamics, calculated with our method and each baseline, represents the trajectory data from the most recent iteration. The datasets used and how they were divided into batches are described below.

\begin{figure}[!t]
\centering
\includegraphics[width=0.9\linewidth]{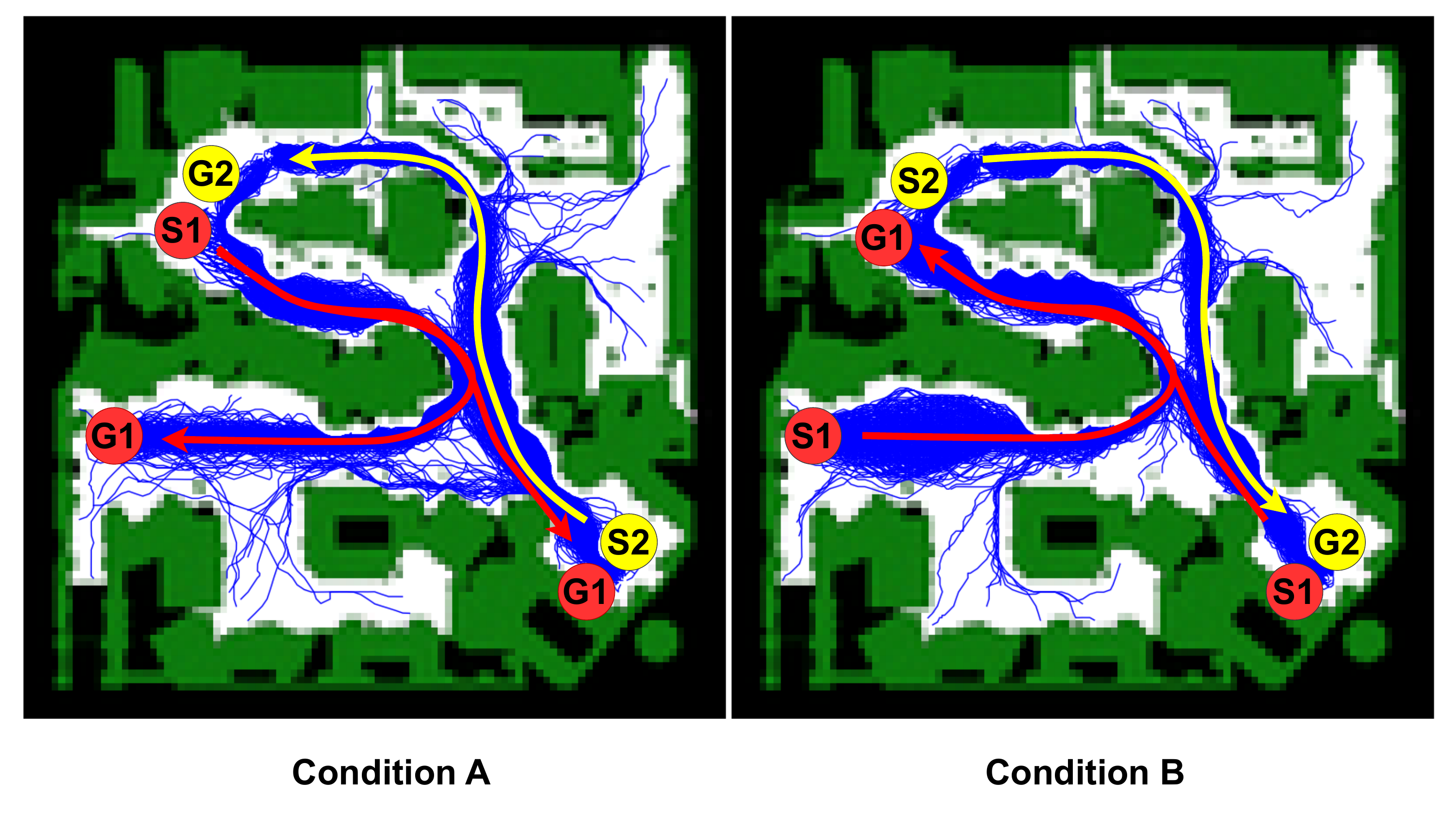}
\vspace*{-2mm}
\caption{The \textit{den520d} synthetic dataset simulates trajectories between positions S1, S2, G1 and G2. The initial flow is shown in Condition A (\textbf{left}). To simulate the flow change, the start (S1, S2) and goal (G1, G2) positions are switched Condition B (\textbf{right}), reversing the dominant flow direction.}
\label{fig:mapf_raw}
\vspace*{-5mm}
\end{figure}

\textbf{\textit{den520d}}: To evaluate the performance of the online update model in scenarios where human flow changes, we introduce a synthetic dataset \textit{den520d}. This dataset is generated using a map from the Multi-Agent Path-Finding (MAPF) Benchmark \cite{stern2019mapf} and features two distinct flow patterns: Condition A and Condition B, see Fig.~\ref{fig:mapf_raw}. We simulate a change in human flow from Condition A to Condition B, where the dominant flow in Condition B is reversed compared to that in Condition A. Furthermore, the \textit{den520d} dataset is used in the downstream robot navigation task (see \cref{sec:nav}) to assess its practical application in flow-aware planning. We discretize the \textit{den520d} obstacle map with \SI{1}{\metre} cell resolution.
In this map, randomized human trajectories are simulated using stochastic optimal control path finding, based on Markov decision processes~\cite{rudenko2020semapp}. In each condition 1000 trajectories are generated. We use the condition B for evaluation in a downstream task (robot motion planning) in Fig.~\ref{fig:mapf_mod}.

\textbf{ATC}: To evaluate the MoD online update in a real-world dataset with scenarios where human motion patterns change during the day, we use the ATC dataset~\cite{brscic2013person}. Using multiple 3D range sensors, the trajectories were collected between 9 am and 9 pm on Wednesdays and Sundays, including 92 days of observations in total. We use the data from the entire first day for evaluation.

In the experiments across both datasets, the grid resolution for MoD is set to \SI{1}{\metre}. The observation rate is downsampled from above \SI{10}{\Hz} to \SI{1}{\Hz}. In our quantitative evaluation, for each condition in the \textit{den520d} or each hour in the ATC, we randomly sample $10\%$ of the data for testing and use the remaining $90\%$ for training.

\subsection{Baselines}
We evaluate several variations of processing the data sequentially using CLiFF-map, and additionally evaluate the prior art STeF-map. The variations in applying CLiFF-map to sequential data include: our online update model (referred to as \textbf{online}), a model built from all observations in iteration $1$ to $k$ (referred to as \textbf{history}) and a model with observations only in iteration $k$ (referred to as \textbf{interval}). When training the CLiFF-map, the convergence precision is set to 1e--5 for both mean shift and EM algorithms, with a maximum iteration count of 100. For online update model, $\eta_{\mathrm{thres}}$ is set to 0.1.

STeF-map~\cite{molina2018modelling} is a spatio-temporal flow map, which models the likelihood of human motion directions on a grid-based map using harmonic functions. STeF-map captures long-term changes of crowd movements over time. Each cell in the grid map contains $\kstef$ temporal models, corresponding to $\kstef$ discretized orientations of people moving through the given cell over time. As suggested in~\cite{molina2018modelling}, $\kstef$ is set to 8 in the experiments and the model orders for training STeF-map, i.e. the number of periodicities, is set to 2. In our experiments, STeF-map is evaluated using the ATC dataset. Due to the characteristics of STeF-map, it can only be applied to data that exhibit meaningful \emph{periodic} time variations and therefore lends itself better to datasets that span multiple days.

\subsection{Metrics}
To quantitatively evaluate the accuracy of modeling human motion patterns (MoD model quality), we use negative log likelihood (NLL) as the evaluation metric. The average NLL is computed for the most recent maps of dynamics over the corresponding test portion of the dataset. A lower NLL value indicates better accuracy. If no dynamics pattern is available for a given location -- a situation more frequent with interval models, which may fail to capture complete motion patterns due to limited data in the recent iteration -- the likelihood is set to a very low value (1e--9). For runtime evaluation, we report the runtime across all iterations, measured in minutes.

\subsection{Robot deployment} \label{sec:deploy}
To verify the method application on a real robot using live input from a human perception stack, we collected motion data of several people in a small indoor environment (\SI{40}{\metre\squared}) in four periods (approx. three minutes of observation each). Each period has a unique motion pattern, e.g. in changing the direction of movement and turns.

\subsection{Downstream task: human-aware motion planning} \label{sec:nav}
In order to evaluate the effects of having more accurate maps of dynamics in the downstream task, we use the \textit{den520d} dataset to demonstrate socially-compliant flow-aware motion planning. With the \textit{den520d} map, an A* algorithm~\cite{hart1968formal} modified to use flow-aware costs~\cite{swaminathan2018down} is used for generating socially-aware paths for agents.

\begin{table}[t]
\vspace{1mm}
    \centering
    \begin{tabular}{lcccc}
     \toprule
        \textbf{Dataset} & \multicolumn{4}{c}{\textbf{Average NLL}} \\
          & Online (ours) & Interval & History & STeF-map \\
        \midrule
        ATC & \textbf{1.18} & 2.61 & 1.23 & 4.38 \\
        \textit{den520d} & \textbf{1.41} & 2.36 & 1.65 & - \\
        \bottomrule
    \end{tabular}
    \caption{Evaluation results on the ATC and \textit{den520d} datasets using average negative log likelihood (NLL) as the metric. Lower values are preferable. Values in bold indicate the best performance.}
    \label{tab:expres}
\vspace*{-2mm}
\end{table}

\section{Results} \label{section-results}

\begin{figure}[t]
\centering
\includegraphics[clip,trim= 0mm 0mm 0mm 0mm,height=27mm]{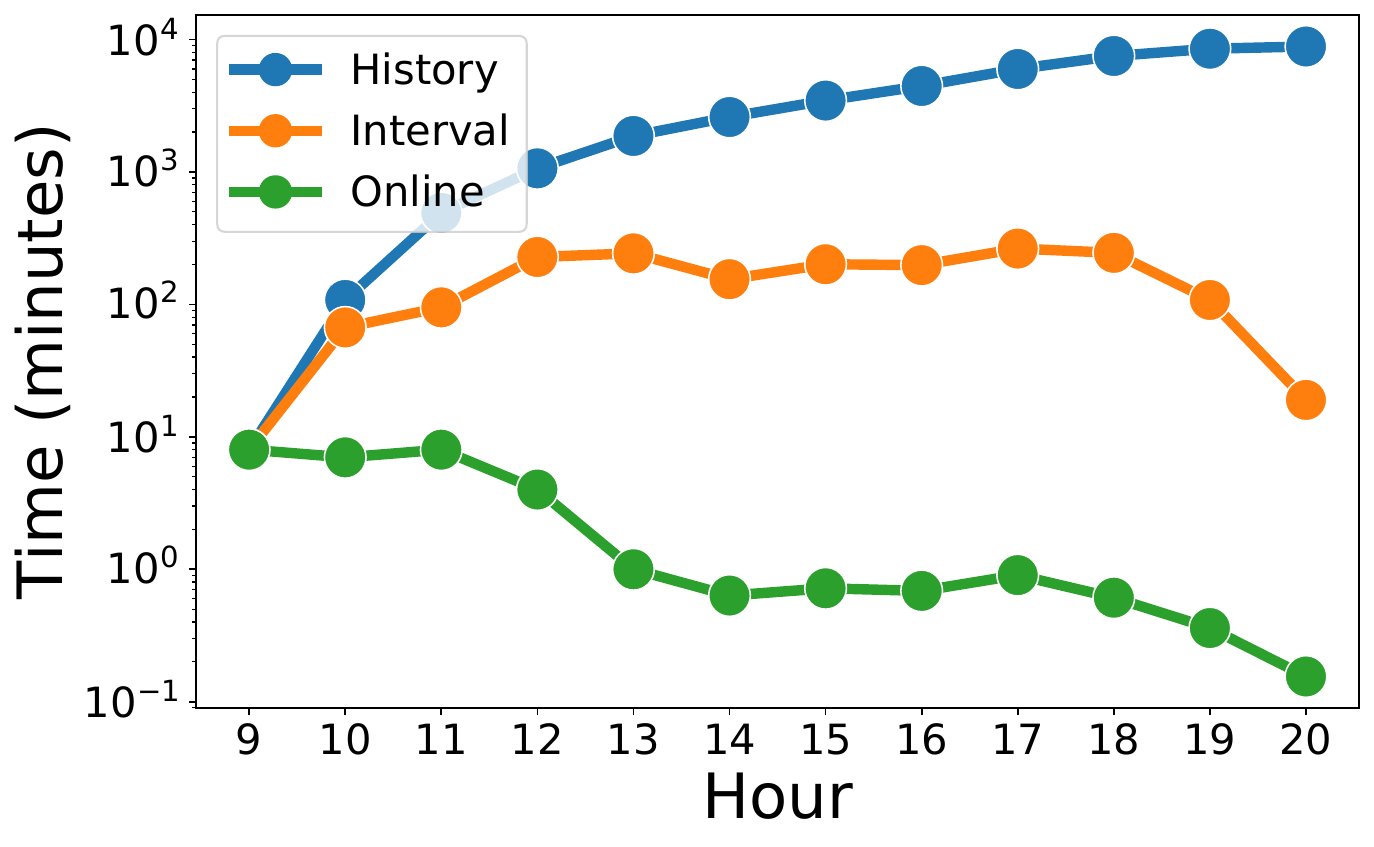}%
\includegraphics[clip,trim= 12mm 0mm 0mm 0mm,height=27mm]{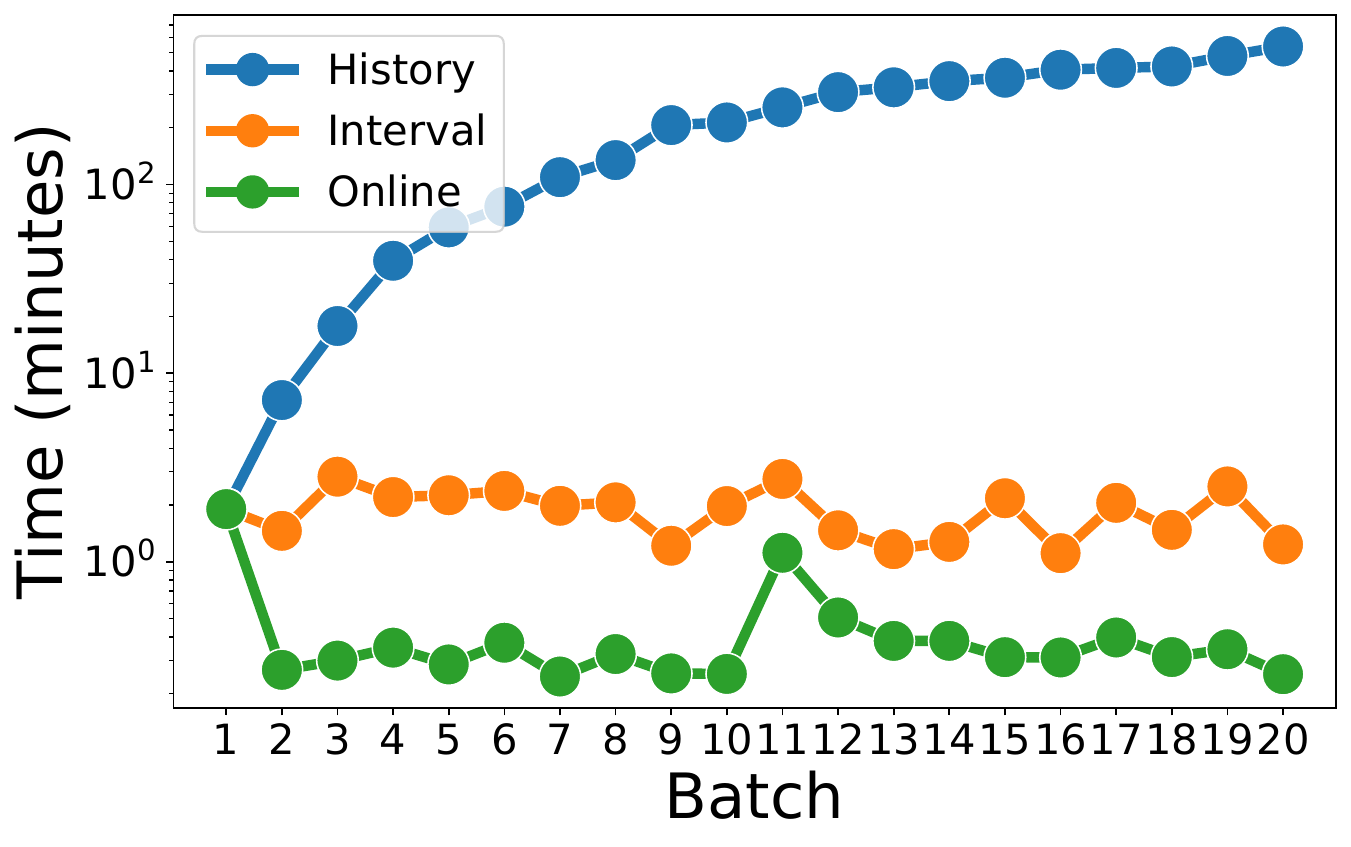}
\caption{Runtime of each iteration in the ATC (\textbf{left}) and \textit{den520d} (\textbf{right}) datasets. In \textit{den520d}, the first 10 batches are in Condition A and the second 10 batches are in Condition B. In both datasets, the online model shows significantly reduced runtime compared to the history and interval models.}
\label{fig:run_time}
\vspace*{-5mm}
\end{figure}

\begin{figure}[t]
\vspace{1mm}
\centering
\includegraphics[width=1.\linewidth]{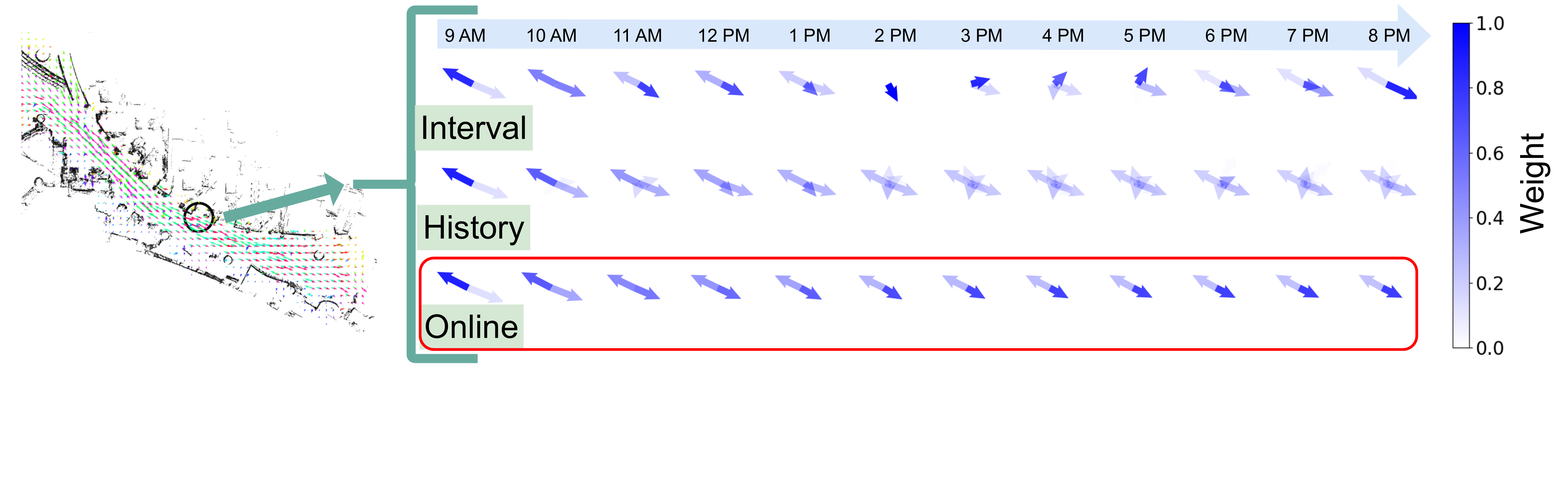}
\vspace{-11mm}
\caption{An example from the east corridor in the ATC dataset. Blue arrows show the mean vectors of SWGMMs, with transparency indicating component weights. The interval model (\textbf{top row}) uses only the last 60 minutes of data, disregarding previously learned patterns (e.g., at 2 PM and 3 PM). The history model (\textbf{second row}), which treats all observations equally, tends to obscure patterns and fails to capture dominant movements effectively. Conversely, the online model (\textbf{third row}) adapts more effectively to changing flows and accurately represents the primary patterns.}
\label{fig:atchour}
\vspace*{-2mm}
\end{figure}

\begin{figure}
\centering
\includegraphics[width=0.48\linewidth]{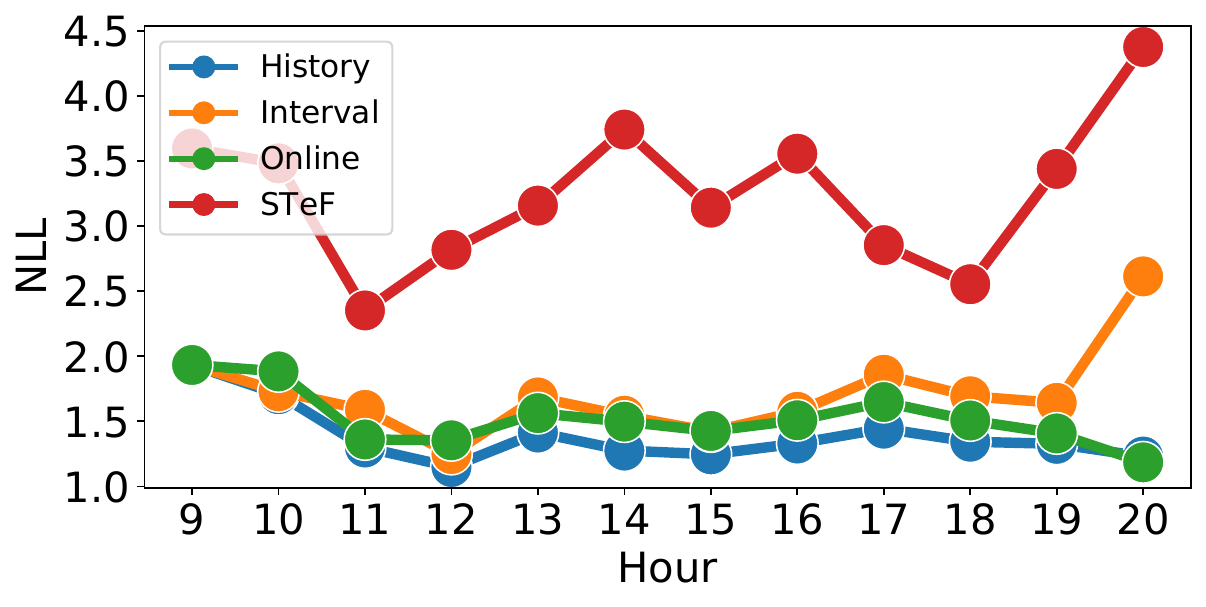}
\includegraphics[width=0.48\linewidth]{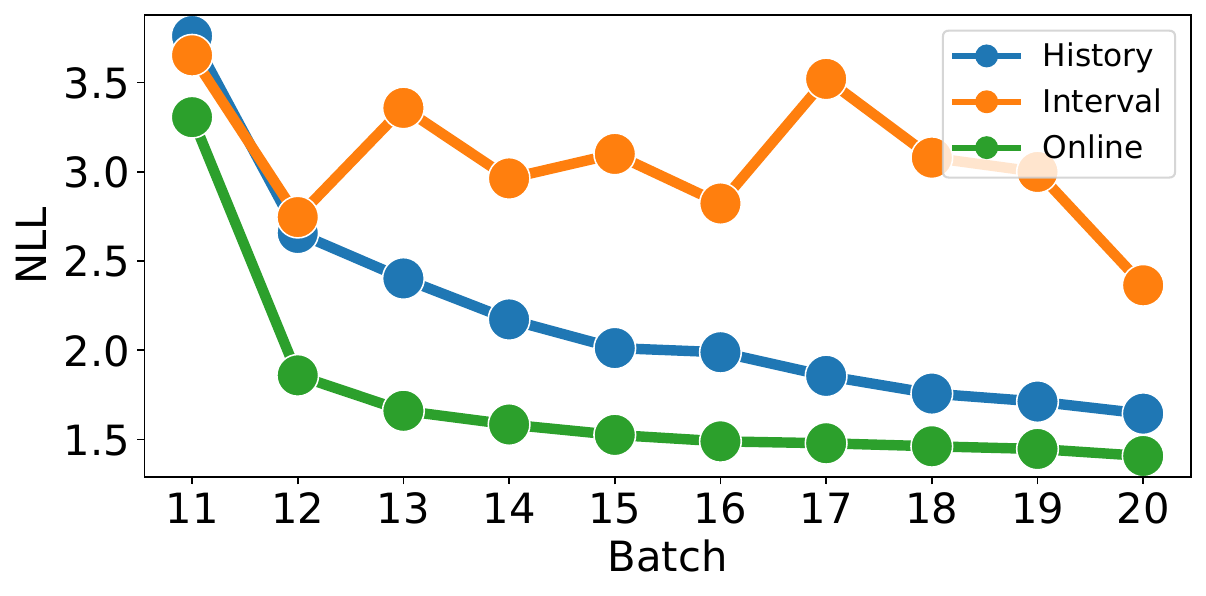}
\caption{\textbf{Left}: NLL evaluation results at different times of a day in the ATC dataset. The STeF-map struggles to accurately capture human motion flow, resulting in poor performance. Notably, the interval model exhibits worse NLL results mostly after 16:00. It struggles to capture the full spectrum of motion patterns and is not able to effectively manage outlier trajectories outside of the main flow. \textbf{Right}: NLL evaluation results over batches in Condition B of the \textit{den520d} dataset. The online model quickly adapts to changes in human flow, achieving lower NLL values in fewer batches.}
\label{fig:batch_nll}
\vspace*{-3mm}
\end{figure}

\begin{figure}[t]
\vspace{1mm}
\centering
\includegraphics[width=0.7\linewidth]{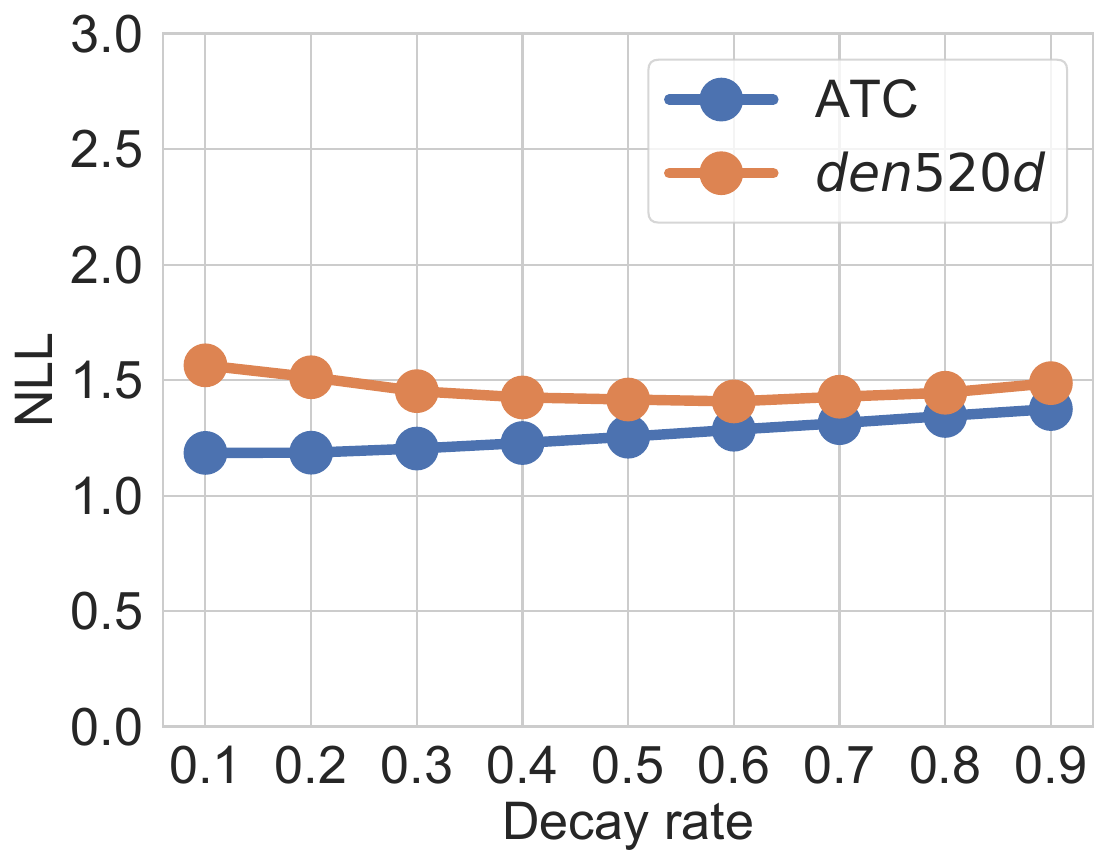}
\caption{Evaluation results of the average NLL for the online update models in the ATC and \textit{den520d} datasets using various decay rates, showing robust performance.}
\label{fig:decay_rate}
\vspace*{-1mm}
\end{figure}

\begin{figure}[t]
\centering
\includegraphics[clip,trim= 0mm 0mm 0mm 0mm,height=35mm]{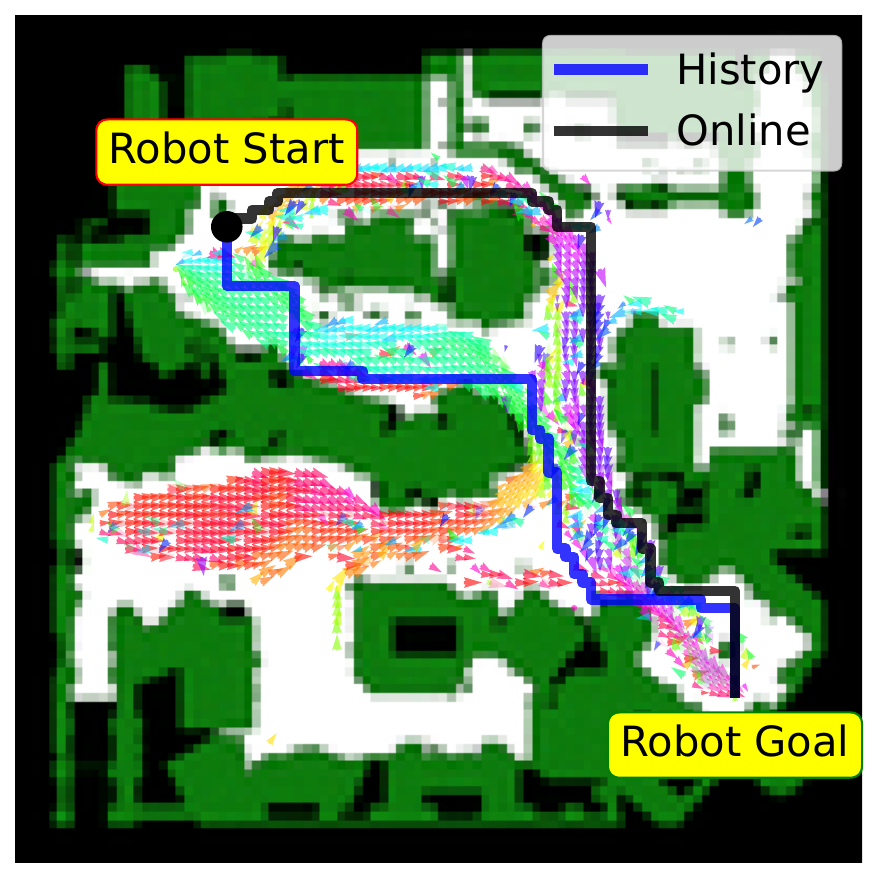}
\includegraphics[clip,trim= 0mm 0mm 38mm 0mm,height=35mm]{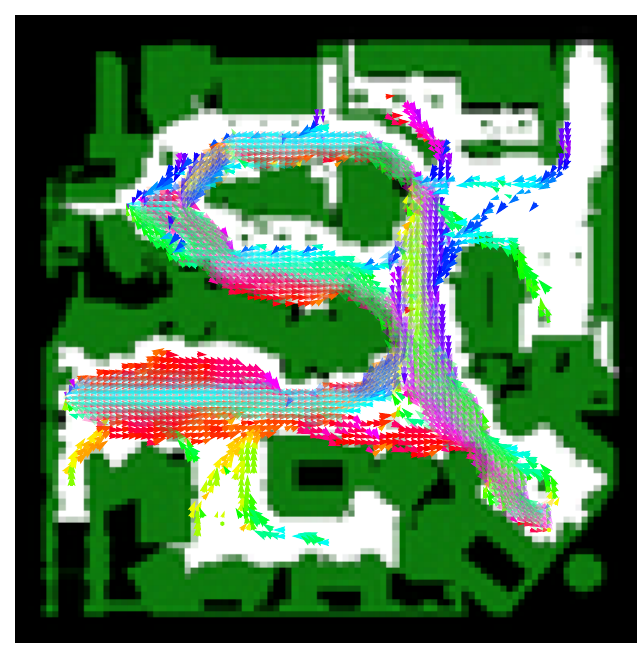}
\includegraphics[clip,trim= 110mm 0mm 0mm 0mm,height=35mm]{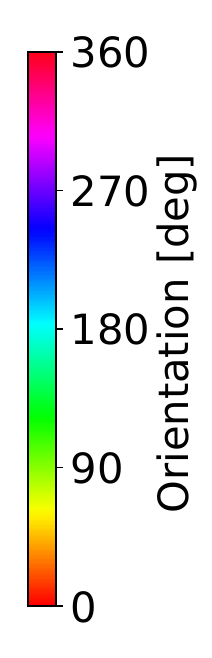}
\caption{Flow-aware robot navigation results (\textbf{left}), using the \textit{den520d} map, where we use an A* algorithm~\cite{hart1968formal} modified to use flow-aware costs~\cite{swaminathan2018down} to generate the agent's path. The MoD of the online model is depicted as an underlying reference for flow direction. Starting from the upper-left corner, the MoD from the history model (\textbf{right}) fails to adapt to flow changes. Consequently, the generated path continues to follow the previous human flow from Condition A, contradicting the current flow in Condition B. In contrast, the online method, which quickly adapts to flow changes, enables the agent to navigate in alignment with the existing human flow.}
\label{fig:mapf_mod}
\vspace*{-3mm}
\end{figure}

\begin{figure}[t]
\vspace{1mm}
\centering
\includegraphics[width=1.\linewidth]{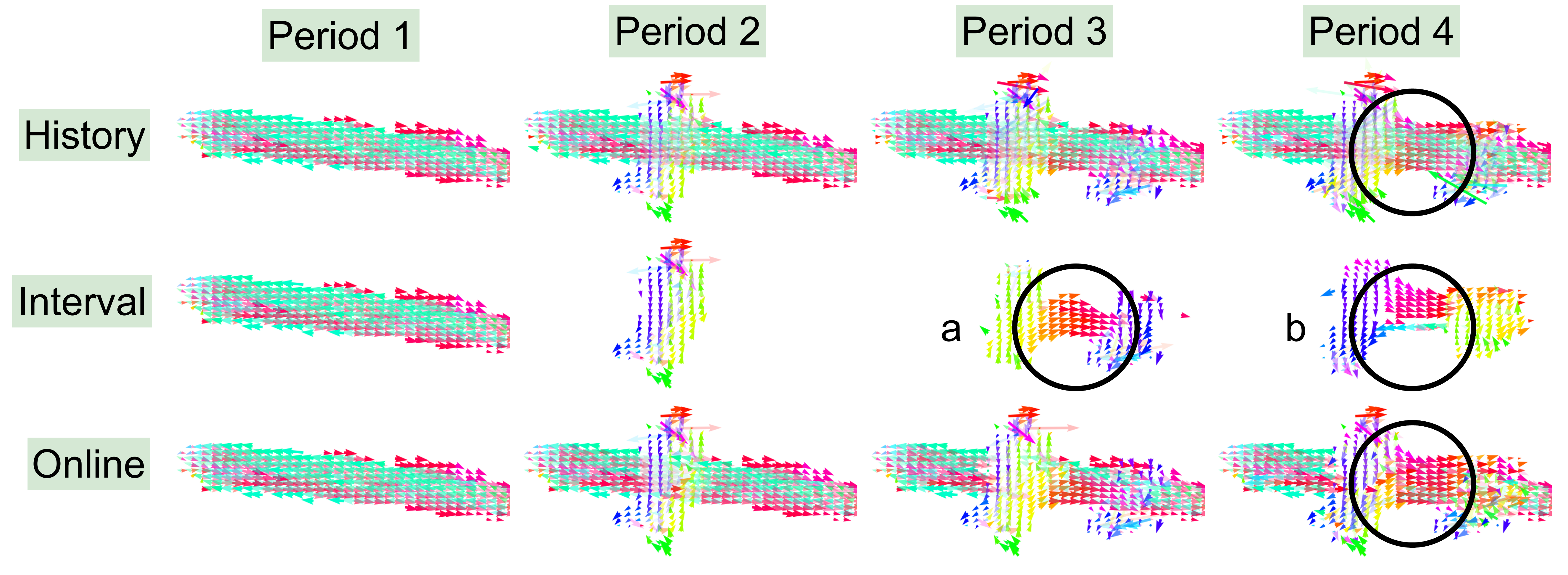}
\caption{Models built from real-world data collected from the robot. This experiment highlights that using all historical data typically converges to the possibility of motion in all directions, whereas the online model reflects the dominance of recent observations, while maintaining the completeness of human flow. For instance, the parts of the map influenced by Periods 3 and 4, highlighted by circles $a$ and $b$), are captured more accurately by the online model (\textbf{bottom row}), compared to the history model (\textbf{top row}).}
\label{fig:realworld}
\vspace*{-3mm}
\end{figure}

In this section, we present the results obtained in the ATC and \textit{den520d} datasets with our approach compared to baselines. The performance evaluation is conducted using both quantitative and qualitative analysis. 


\subsection{Quantitative evaluation}
\subsubsection{Runtime}
The first experiment evaluates the runtime of the online method, compared to the history model and interval model, in both datasets.
We use a desktop computer with an Intel i9-12900K processor running Ubuntu 20.04. As shown in \cref{fig:run_time}, the online update model exhibits significantly reduced computation time compared to the interval and history models. In the initial iteration, the MoD is empty, and all three models are built from scratch using all available observations. Subsequently, the online update model computes the sufficient statistics with $N_k$ observations in each iteration. In contrast, the interval model runs the EM algorithm to converge (to 1e--5) using $N_k$ observations, and the history model processes cumulative, $\Sigma_{k=1}^{K}N_k$ observations. On average, over all iterations, the online model requires 0.1\% of the time used by the history model in ATC and 0.2\% in \textit{den520d} dataset.

\subsubsection{Accuracy}
The second experiment evaluates the accuracy of modeling human motion patterns. We present the average negative log likelihood (NLL) values, with lower values indicating better modeling accuracy. These values are optimized using the decay rate parameter. \cref{tab:expres} shows how well the MoDs reflect changes in human dynamics, using trajectory data from Condition B of the \textit{den520d} dataset and the last hour of the ATC dataset. 

Compared to the STeF-map, which discretizes orientations into eight bins, the CLiFF-map provides a continuous probabilistic representation of velocities, achieving better accuracy in modeling human motion patterns. Among variations of CLiFF-map construction, the online update model achieves the best accuracy when the motion patterns change, as it accurately captures these changes while maintaining flow completeness. In contrast, the history model fails to adapt to changes in human flow, and the interval model suffers from insufficient data within its given interval, leading to several parts of the map having no flow representation, failing to preserve flow completeness. In the ATC dataset, the most notable human motion pattern changes occur in the east corridor. A detailed view of this area is provided in \cref{fig:atchour}, where the online update model's adaptations are shown, transitioning from predominantly upward movement to primarily downward movement.

\cref{fig:batch_nll} shows the evaluation results for each hour in the ATC dataset and each batch in Condition B of the \textit{den520d} dataset. In the ATC dataset, STeF-map struggles to accurately capture human motion flow, resulting in poor performance. The interval model exhibits worse NLL results, mostly after 16:00, due to its failure to capture the full spectrum of motion patterns. During the middle of the day, when hourly motion patterns are closer to the general motion pattern, the history model performs better by leveraging accumulated data over time. However, without prioritizing recent observations, the history model fails to accurately reflect the most recent motion patterns. In contrast, the online model, with its ability to effectively adapt to changes, achieves better performance in the last hour of the ATC dataset. In the \textit{den520d} dataset, which exhibits a clear flow change from Condition A to Condition B, the history model takes more batches to reflect the flow change, while the online model quickly adapts to changes and achieves lower NLL values in fewer batches.

\noindent\textbf{Parameter analysis}: We evaluate the average NLL value with different decay factor values $\lambda$, varying from 0.1 to 0.9, in both ATC and \textit{den520d} datasets. The evaluation results are shown in \cref{fig:decay_rate}. We observe robust performance across different decay rate values.

\subsection{Qualitative evaluation}
\cref{fig:mapf_mod} shows the CLiFF-maps built with online and history models in Condition B of the \textit{den520d} dataset. When human flow shifts from Condition A to B, the online model quickly accommodates the change, while the history model preserves both flow directions and struggles to adapt. This inaccuracy in the MoD built by the history model affects the downstream task of flow-aware robot navigation; as shown in \cref{fig:mapf_mod} (left), the path generated using the history model continues to align with the previous human flow from Condition A, which contradicts the current flow in Condition B, resulting in increased collisions. Conversely, the online method quickly adapts to flow changes, enabling the agent to navigate in a manner that aligns with the current flow pattern.

In the real-robot experiment, we use live input from a human perception stack, with each period showing a unique motion pattern. To demonstrate the runtime bottleneck of baseline methods for real-time applications, we compare runtime on the most recent batch of observations. The history and interval models require \SI{990}{\second} and \SI{29}{\second}, respectively, making them impractical for real-time deployment. In contrast, the proposed online model runs in \SI{5}{\second}, using only 0.5\% of the history model's time, making it more suitable for real-time operation. When comparing the quality of the generated MoD, \cref{fig:realworld} shows that in the last period, while the history model contains motion possibilities in all directions, and the interval model completely forgets previous information and leaves parts of the map of dynamics blank, the online model effectively maintains the dominance of recent observations while keeping the completeness of human flow.

\section{Conclusion} \label{section-conclusions}
In this work we propose a method for online updates of the CLiFF-map of dynamics representation, enabling it to quickly adapt to human flow changes. We use the stochastic expectation maximization algorithm to update the existing representation when new observations are collected. The method is evaluated using both synthetic and real-world datasets, and compared with three baselines. Additionally, we evaluate the method in a downstream task of socially-compliant, flow-aware motion planning. The results demonstrate that our online version can efficiently and accurately accommodate changes in human motion patterns. Furthermore, it is significantly faster than the baselines and avoids the large memory consumption associated with storing the entire dataset of historic observations.

The proposed method has limitations. As a result of our experiments with various modes of tracking the dynamics observation history, we notice that maintaining an up-to-date MoD is a challenging problem due to the need to balance the historic patterns vs. deciding when the data is outdated \cite{Biber2005DynamicMF}. While the proposed online update method significantly improves runtime, making it practical for real-time robotics applications, it may not always deliver optimal performance. In situations where dynamic patterns and flow of motion in the environment are complex, historical models can achieve better accuracy as they are better equipped to handle outlier trajectories that deviate from the main flow pattern. Conversely, when the human motion flow is more laminar and there is sufficient current observational data to accurately represent human flow, the interval model may capture the most accurate patterns of human motion and can serve as a benchmark for evaluation. In these instances, the online update methods may perform less effectively compared to the interval model.

In future work we intend to use the online update method as an initial building block to enable efficient robot exploration in dynamic environments.










\printbibliography
\end{document}